\documentclass{article} % For LaTeX2e
\usepackage[final]{nips_2017}
\usepackage{hyperref}
\usepackage{url}
\usepackage{amsmath}
\usepackage{amssymb}
\usepackage[ruled]{algorithm2e}
\usepackage{graphicx}
\usepackage{wrapfig}
\usepackage{subcaption}

\DeclareMathOperator{\E}{\mathbb{E}}
\DeclareMathOperator*{\argmax}{arg\,max}

\title{Forward-Backward Reinforcement Learning}

\author{
Ashley D. Edwards\thanks{Work done as an intern at Google Brain}\thanks{Correspondence to: aedwards8@gatech.edu}\textsuperscript{\;\; $\bowtie$ *}, Laura Downs\textsuperscript{$\ddagger$}, James C. Davidson\textsuperscript{$\ddagger$} \\
\textsuperscript{$\bowtie$}Georgia Institute of Technology \\
\textsuperscript{$\ddagger$}Google Brain \\
\vspace{-3em}
}

\begin{document}

\maketitle

\begin{abstract}
% Reinforcement learning is often formulated with the agent blind to the task reward of the environment.  However, for many problems, the reward can be decoupled from the transition model. For many sparse reward problems, including goal-directed tasks such as point-to-point navigation, pick and place manipulation, assembly, etc., endowing the agent with knowledge of the reward function both feasible and practical for learning generalizable behavior.
% Added this to the intro 
Goals for reinforcement learning problems are typically defined through hand-specified rewards. To design such problems, developers of learning algorithms must inherently be aware of what the task goals are, yet we often require agents to discover them on their own without any supervision beyond these sparse rewards. While much of the power of reinforcement learning derives from the concept that agents can learn with little guidance, this requirement greatly burdens the training process. If we relax this one restriction and endow the agent with knowledge of the reward function, and in particular of the goal, we can leverage backwards induction to accelerate training. To achieve this, we propose training a model to learn to take imagined reversal steps from known goal states. Rather than training an agent exclusively to determine how to reach a goal while moving forwards in time, our approach travels backwards to jointly predict how we got there. We evaluate our work in Gridworld and Towers of Hanoi and empirically demonstrate that it yields better performance than standard DDQN. 
\end{abstract}

\section{Introduction}
Reinforcement Learning (RL) problems are often formulated with the agent blind to the task reward of the environment.  However, for many sparse reward problems, including goal-directed tasks such as point-to-point navigation, pick-and-place manipulation, assembly, etc., endowing the agent with knowledge of the reward function is both feasible and practical for learning generalizable behavior. In general, developers of these problems often know what the task goals are, but not necessarily how to solve them. In this paper, we will describe how we can leverage our knowledge of goals to enable learning of behaviors in these regions before the agent even reaches them. This formulation may be easier to solve than approaches that initialize learning from the start alone. For example, if we know the desired location, pose, or configuration of a task, then we can reverse the actions that brought us there, rather than forcing the agent to solve these difficult problems solely through random discovery.

In this paper, we introduce Forward-Backward Reinforcement Learning (FBRL), which introduces backward induction, to enable our agent to reason backwards in time.  Through an iterative process, we both explore forwards from the start position and backwards from the target/goal.  To achieve this we introduce a learned backwards dynamics model to explore in reverse from known goal states and update values within this local neighborhood. This has the effect of ``spreading out" sparse rewards so that they are easier to discover, thus accelerating the learning process. 

Standard model-based approaches aim to reduce the amount of experience necessary to learn good policies by imagining steps~\emph{forward} and using these hallucinated events to augment the training data. However, there is no guarantee that the projected states will lead to the goal, so these roll-outs may be inadequate.  The ability to predict the result of an action does not necessarily provide guidance about which actions lead to the goal. In contrast, FBRL takes a more guided approach since, given an accurate model, we have confidence that each state visited in a backwards step has a path to the goal.

In the rest of the paper, we will describe the relevant background and related works. We will then formally introduce FBRL, followed by an empirical section in which we evaluate our approach in Gridworld and Towers of Hanoi, and show that it yields better results than standard Deep Double Q-Learning (DDQN)~\citep{van2016deep}. Finally, we will conclude with discussions for future work.
% We consider directed and undirected learning from the goal states.  In undirected learning, we take random reverse actions from the goal states.  In directed learning, we can either take steps designed to improve our knowledge of the value function or designed to seek out the initial state.

\section{Background}
Reinforcement Learning (RL) problems are specified through a Markov Decision Process (MDP) $\langle S, A, R, T \rangle$~\citep{sutton1998reinforcement}. Here, $s \in S$ describes the states in the environment, $a \in A$ defines the actions the agent can take, $r \in R(s)$ refers to the rewards an agent receives within state $s$, and $T(s, a, s')$ is a transition model that specifies the probability of entering state $s'$ after taking action $a$ in $s$. A policy $\pi$ estimates the probability of taking action $a$ in state $s$, and we are typically interested in learning an optimal policy that maximizes the expected long-term discounted return. Model-free approaches do not have access to $T$, and rather learn an action-value function $Q(s, a)$ that predicts the return after experiencing samples $\langle s, a, s', r \rangle$ in the environment:
\begin{equation}
\label{equation:TD}
L(\theta) = \E_{(s, a, s', r)\sim D}[r + \gamma \max_{a'}Q(s', a') - Q(s, a)]
\end{equation}

Here, $D$ is a replay buffer that stores experiences~\citep{mnih2015human}. This loss aims to minimize the TD-error, or the difference between the expected return and current prediction. 

Learning Q-values often requires a large quantity of samples. Rather than directly experiencing the states, an alternative method is to jointly use model-based planning to predict values. DYNA-Q~\citep{sutton1990integrated} makes updates to values by using imagined experiences. In this case, the parameters $\langle s, a, r, s'\rangle$ from Equation~\ref{equation:TD} may also be obtained from imagined experiences.

% As in \cite{DBLP:journals/corr/HenaffWL17}, we use 1-hot encoding to select actions from a discrete action space, but we select actions rather than states.  Rather than performing a gradient search through the reverse dynamics model, we use the model to explore the space stochastically.

\section{Related Work}
When we have access to the true dynamics model, purely model-based approaches such as dynamic programming can be used to compute values over all states~\citep{sutton1998reinforcement}. Though when the state space is large or continuous, it may be intractable to iterate over the entire state-space. Q-Learning is a model-free approach and updates values in an online manner by directly visiting states, and function approximation techniques such as Deep Q-Learning enable generalizing to unseen ones~\citep{mnih2015human}. Hybrid approaches that combine model-based and model-free information can also be used. 
 DYNA-Q~\citep{sutton1990integrated}, for example, was an early approach that used imagined roll-outs to update the Q-values as if they had been experienced in the true environment. There are more recent approaches as well, for example NAF~\citep{gu2016continuous} and I2A~\citep{weber2017imagination}. But these approaches only use forward imagination.  

A similar approach to our own does value iteration in reverse~\citep{zang2007horizon}, but this is a purely model-based approach, and it does not learn a reverse model. A related approach performs bidirectional search from the start and goal~\citep{baldassarre2003forward}, but that work learns values only, whereas we aim to learn action-values. Another comparable work solves problems by using a reverse curriculum near goal states~\citep{florensa2017reverse}. However, that approach assumes the agent can be initialized near the goal. We do not make this assumption, as knowing what the goal state is does not mean that we know how to get to it.  

Many works have used domain knowledge to help speed up learning, for example through reward shaping~\citep{ng1999policy}. Another approach is to more efficiently use the experiences from the replay buffer. Prioritized experience replay~\citep{schaul2015prioritized} aims to replay samples that have high TD-error. Hindsight experience replay treats each state in an environment as a potential goal so that the system can learn even when it fails to reach the desired target.

The concept of using reverse dynamics is similar to inverse dynamics~\citep{agrawal2016learning, pathak2017curiosity}. In those approaches, a system predicts the dynamics that yielded a transition between two states. In our approach, we use the state and action to predict the previous state. The purpose of this function is to reverse an action and use this unraveling to learn values near the goal.  

\section{Approach}
\begin{wrapfigure}{R}{0.5\textwidth}
    \begin{minipage}{0.5\textwidth}
        \begin{algorithm}[H]
          \SetAlgoLined
          \caption{Forward-Backward RL}
          \BlankLine
         \While{training}{
         /* Forward step */
         
         Take step from $\pi$ and update $D$
          
          $d \sim D$ 
          
          Train $b(\cdot)$, $Q(\cdot)$ with $d$

          $\widehat{s}_{t+1} \sim G$ 
          
          /* Backward step */
          
          \For{imagination steps}{
            $a_t \gets $ random or greedy
            
            $r_t \gets R(\widehat{s}_{t+1})$
            
            $\widehat{s_t} \gets \widehat{s}_{t+1} - b(\widehat{s}_{t+1}, a_t)$
            
            $D.append(\widehat{s}_t, a_t, r_t, \widehat{s}_{t+1})$
            
            $\widehat{s}_{t+1} \gets \widehat{s}_t$
          }
          
         }
         \label{algorithm:fb}
        \end{algorithm}
   \end{minipage}
\end{wrapfigure}
We now introduce our approach, Forward-Backward Reinforcement Learning (FBRL). In this work, we utilize both imagined and real experiences to learn values. A~\emph{forward} step uses samples of real experiences originating from the start state to update Q-values, and a~\emph{backward} step uses imagined states that are asynchronously predicted in reverse from known goal states. We hypothesize that this approach will improve our model of values in the vicinity of the goal, and thus expedite learning. We now describe the preliminaries for our approach.

\subsection{Preliminaries}
We specify FBRL problems through a modified MDP $\langle S, A, R,  G\rangle$. As before, $s \in S$ corresponds to the states in the environment, $a \in A$ are the actions the agent can take, and $R(s)$ represents the rewards an agent receives in $s$. We assume that $R$ does not distinguish between real and imagined inputs and can be queried at any time. Finally, $g \sim G$ is a distribution of goal states from which we can sample uniformly.

\subsection{Backwards model}
We aim to learn a backward transition model that captures what happens if we undo an action in a state. We use a tuple of experience $\langle s_t, a_t, r_t, s_{t+1} \rangle \sim D$ to learn the model.
Rather than predicting the previous state directly, we aim to learn the difference between the two:
$\Delta = s_{t+1} - s_t$. This allows the model to learn how states will change, rather than absolute positional information.
%The forward and backward model are estimating the same change of state given different inputs.
It reduces the expected range of output values and generally centers them around zero, resulting in a more stable estimate.
This formulation is appropriate since we are using states from the start of the problem to learn the backwards model, which is used near goal states that will initially have little training data.

%\begin{equation}
%b(s_{t+1}, a_t) = f(s_t, a_t) = s_{t+1} - s_t = \Delta
%\end{equation}

The backwards model is a neural network that is trained to predict $\Delta$, where $b(s_{t+1}, a_t) \rightarrow \widehat{\Delta}$. Now, we can predict the previous state as $\widehat{s}_t = s_{t+1}  - b(s_{t+1}, a_t)$. The loss for the backward model then is:
$\mathcal{L}_{\theta_{b}} = \lVert \Delta - b(s_{t+1}, a_t) \rVert$, where $\lVert \cdot \rVert$ denotes a Huber loss.

In some environments, it may be impossible to learn an accurate deterministic backward model, even if the problem has deterministic actions. For example, if an agent is next to a wall, we might not know if it previously bumped into the wall or if it took a step towards it. Additionally, for discrete-valued problems, it may be difficult to learn a network that can predict discrete values. These issues are compounded further in stochastic settings. To address this we formulate the problem using a variational approach. If we know the distribution over $\Delta$, then we can predict a distribution over potential outcomes. In this formulation, $\widehat{\Delta}$ will represent a probability distribution for each state variable that can be trained using a cross-entropy loss from the true distribution.

% In order to regularize the dynamics, we use the alternative model with fixed weights:
% \begin{align*}
%     \mathcal{L}_{\theta_{f}} &= \lVert \Delta - f(\widehat{s_t}, a_t; \theta_b) \rVert \\
%     \mathcal{L}_{\theta_{b}} &= \lVert \Delta - b(\widehat{s}_{t+1}, a_t; \theta_f) \rVert
% \end{align*}

%We also use target networks for the forward and backward networks and update them periodically to ensure a stable target.
\subsection{Action sampling}
Another important consideration is how to sample actions that lead to useful updates. Our approach either randomly samples actions or uses a more greedy step that aims to direct the roll-outs towards the start by moving to states with high Q-values: $\argmax_{a_t} Q(\widehat{s}_t, a_t)$.  

\subsection{Backwards Imagination}
Algorithm~\ref{algorithm:fb} shows the pseudo-code for our approach. In the~\emph{forward step}, we train the agent using experiences from the replay buffer, according to whichever learning paradigm we choose. In this work, we use DDQN. We additionally use real experiences to update the backward model.

The~\emph{backward step} takes place asynchronously. During this process, we use backward imagination for a limited amount of steps. Starting from the goal state, the approach samples an action, uses the model to imagine backwards, and then repeats the process from the resulting state. These imagined experiences are used to augment the replay buffer.

It is important to note that initially the backwards model is unlikely to accurately predict the true dynamics model. The model starts by being trained on experience near the starting region. Often, the portion of the dynamics model exercised outside of this initial region will vary significantly, especially near the goal. For example, consider a maze for navigation task where the maze beyond is unknown or the difference in dynamics for a humanoid lying down versus standing up.

While the model may start out being inaccurate, it provides a constantly improving signal that helps formulate the value function, which is then used to guide exploration. In this way, it acts like an intrinsic reward to provide a predicted direction for exploration for the model. Consider again the navigation problem, where the model in the immediate region will learn a factored representation for locomotion, but cannot predict the walls of the maze further away.  The hallucinated experience will likely predict movement through walls. While this is is inaccurate, it does provide a shape for the value function that will encourage traveling towards the goal until a wall is discovered. Once discovered, the model will update and the value function will shift to anticipate the presence of the wall. As training progresses, the system will capture larger regional dynamics and start to predict potential global dynamics, e.g., presence of walls beyond what has been directly observed.  As the system approaches the goal, the backward model will converge to the real model.

% Are backward steps always random or do we do $\argmin_{a_t} Q(\widehat{s}_{t+1}, a_t)$ to get the best improvement on the value function?
%We also use target networks for the forward and backward networks and update them periodically to ensure a stable target.
% \subsection{Model Regularization}
% To ensure that the backward model learns quickly and effectively, we regularize learning we introduce a forward model in addition to the backwards model and regularize each by the other. The forward model is formulated to predict next predicted stat given the current action: $f(s_t, a_t) \rightarrow s_t + \Delta$. This ensures the predicted delta function that is learned is invertible through both functions, which we observed stabilized training significantly.

\section{Experiments}
\begin{figure}[htb]
  \centering
  \begin{subfigure}{.4\linewidth}
    \centering
    \includegraphics[width=.45\linewidth]{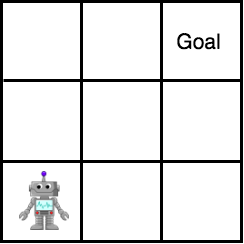}
  \end{subfigure}
  \begin{subfigure}{.4\linewidth}
    \centering
    \includegraphics[width=.65\linewidth]{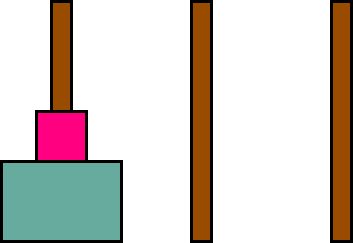}
  \end{subfigure}  
  \caption[]{Gridworld and Towers of Hanoi environments.}  
  \label{fig:environments}
\end{figure}
The purpose of our experiments is to demonstrate that FBRL can significantly speed up learning in environments with sparse rewards. We evaluate our approach in Gridworld and Towers of Hanoi, illustrated in Figure \ref{fig:environments}. For comparison we formulate FBRL by augmented DDQN, which we compare against a standard DDQN baseline. 

\subsection{Gridworld}
\begin{figure}[htb]
  \centering
  \begin{subfigure}{.49\linewidth}
    \centering
    \includegraphics[width=\linewidth]{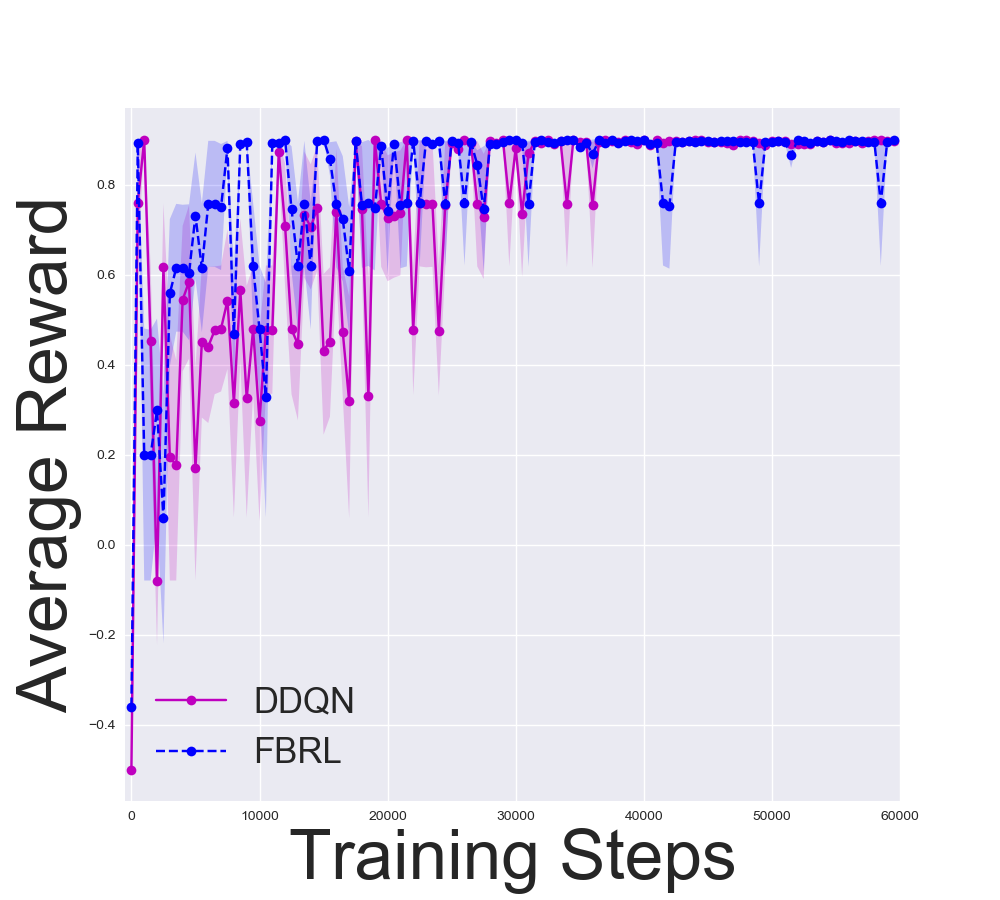}
  \end{subfigure}
  \begin{subfigure}{.49\linewidth}
    \centering
    \includegraphics[width=\linewidth]{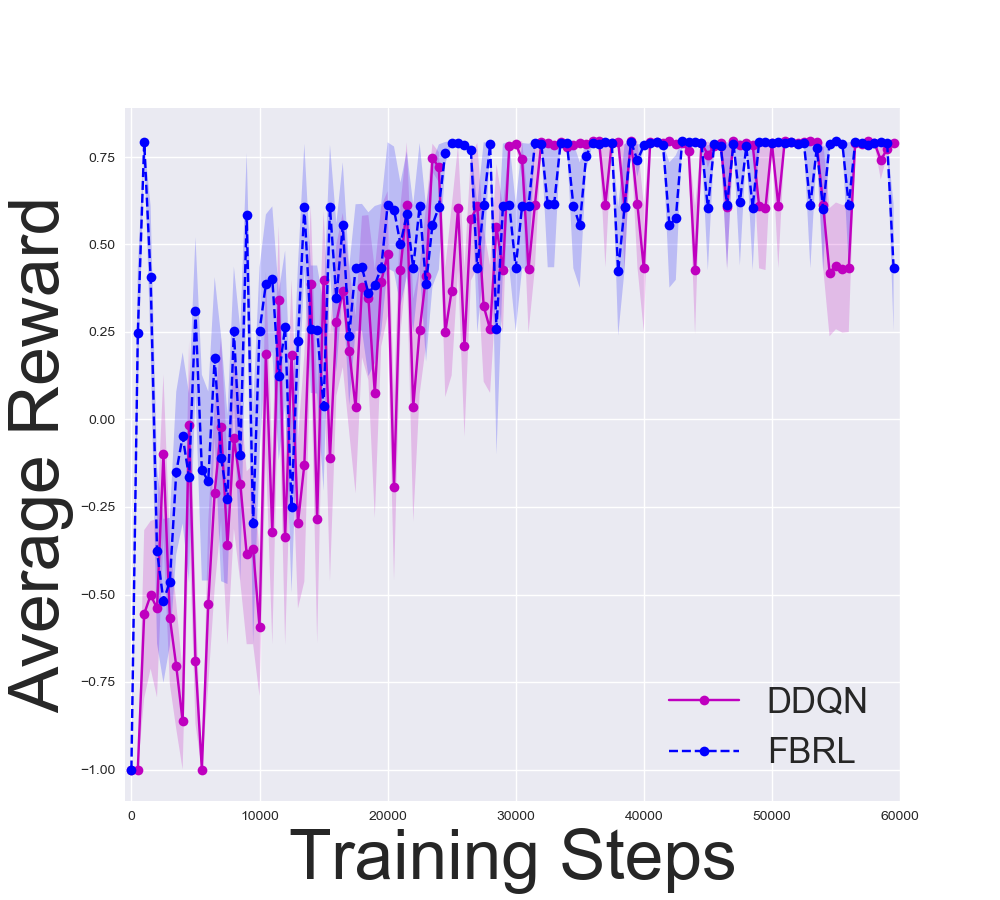}
  \end{subfigure}  
  \\
    \begin{subfigure}{.49\linewidth}
    \centering
    \includegraphics[width=\linewidth]{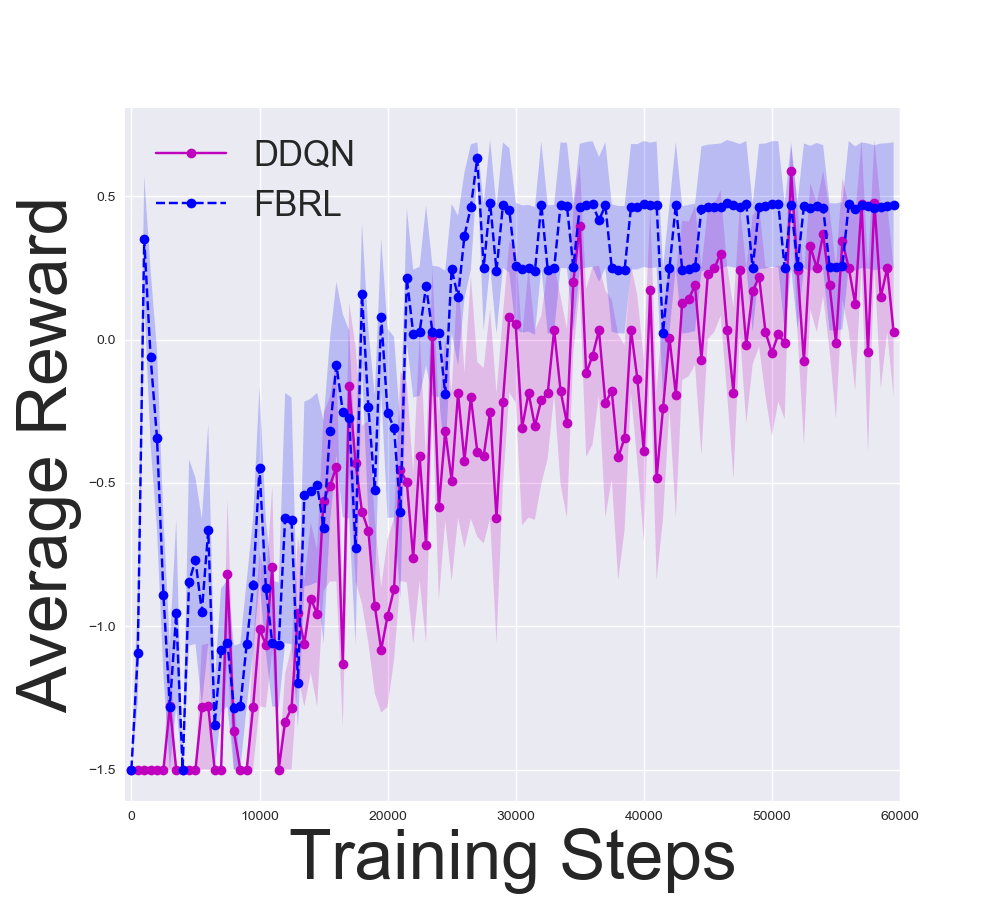}
  \end{subfigure}  
      \begin{subfigure}{.49\linewidth}
    \centering
    \includegraphics[width=\linewidth]{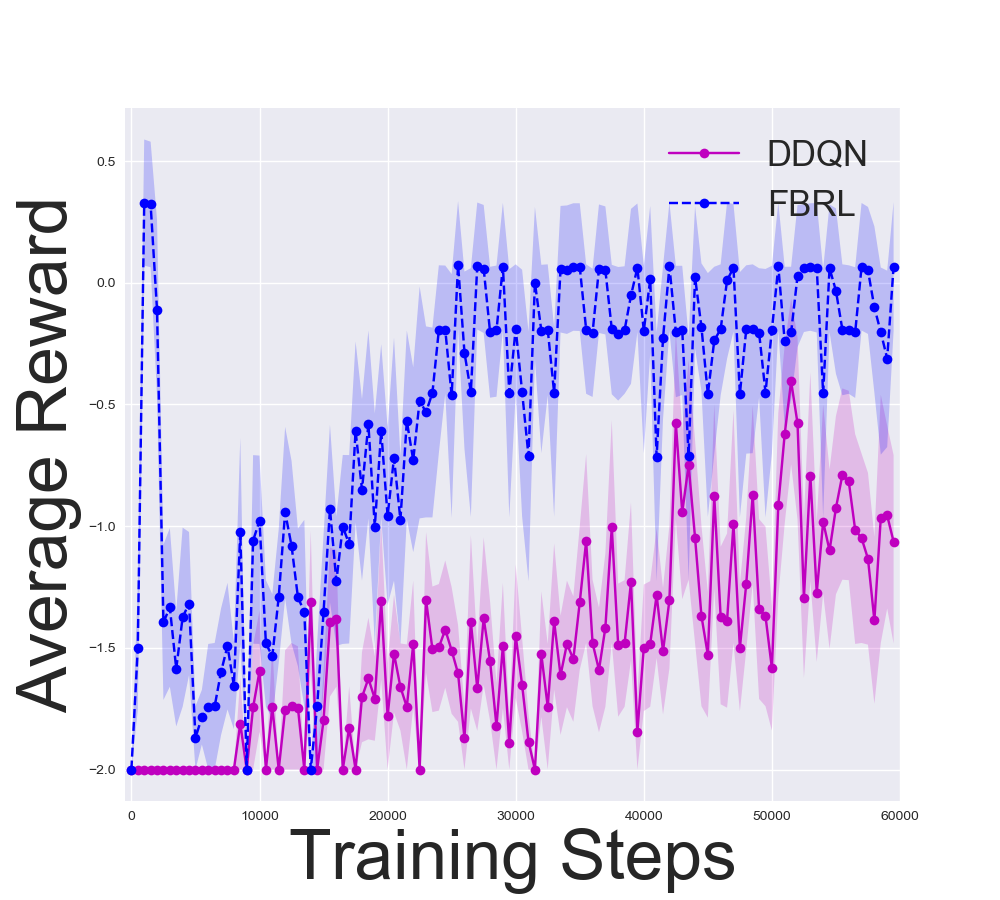}
  \end{subfigure}  
  \caption[]{Results for Gridworld where $n=5,10,15,20$. We use a fixed horizon of $50, 100, 150,200$ steps, respectively. The results are averaged over $10$ trials.}  
  \label{fig:results}
\end{figure}
We first evaluate our approach in an $n$x$n$ Gridworld. We use this environment as it allows us to easily show the benefits of our approach as the reward becomes more sparse. The agent's actions are to move up, down, left, and right by a single unit, and its state consists of its cartesian coordinates. The agent is initialized in the bottom left corner of the grid, and receives a reward of $1$ when it reaches the top right. It receives a step cost of $-.01$ per time-step. The inputs to the backward model are $x,y$ and it must learn to predict $\Delta_x, \Delta_y$. The model architecture is a fully-connected network with $100$ outputs followed by RELU, followed by another fully-connected network with $2$ outputs, one for each state dimension. For FBRL, we used $10$ steps of imagination with $1$ asynchronous stream.

Figure~\ref{fig:results} shows the results for running different size gridworlds. The results show that as we increase the size of the grid, i.e., as the goal gets further away, there is a clear advantage for using reverse imagination. The gap between the performance of DDQN compared to FBRL increases as the size gets larger. This suggests the approach is better suited for longer horizon sparse reward environments--but still does not degrade performance for short horizon tasks.

\subsection{Towers of Hanoi}
\begin{figure}[htb]
  \centering
  \begin{subfigure}{.49\linewidth}
    \centering
    \includegraphics[width=\linewidth]{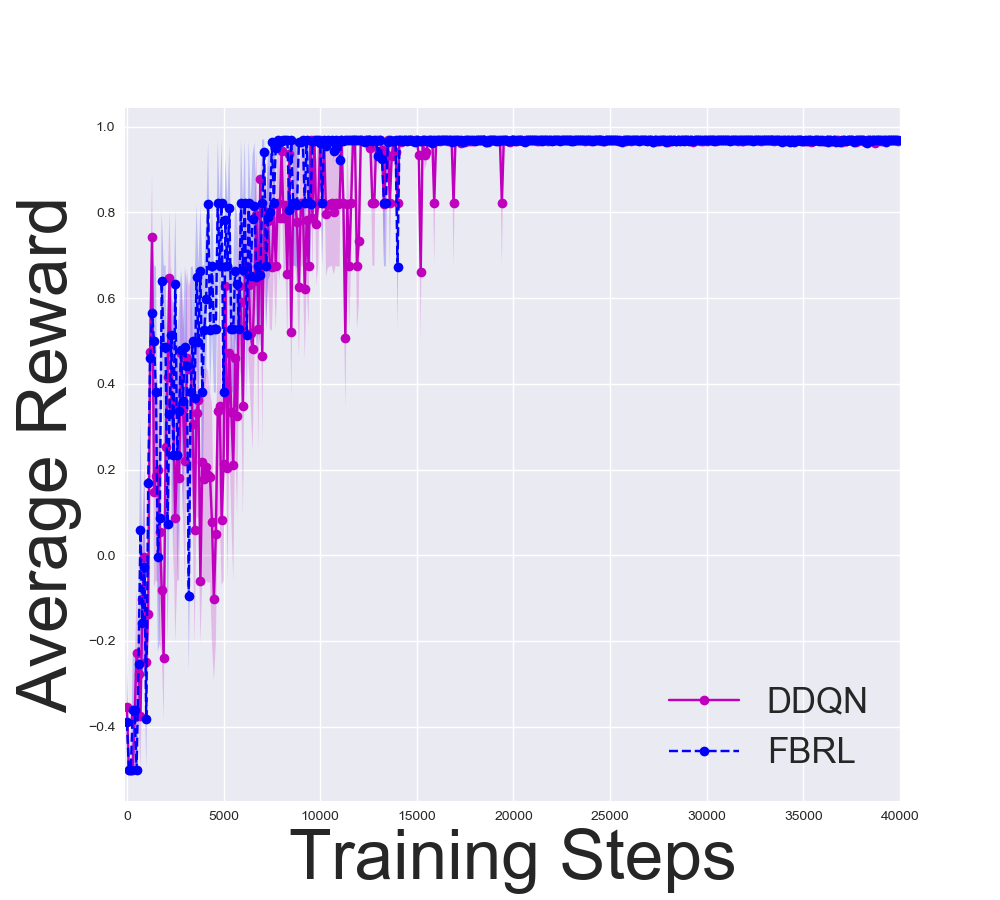}
  \end{subfigure}
  \begin{subfigure}{.49\linewidth}
    \centering
    \includegraphics[width=\linewidth]{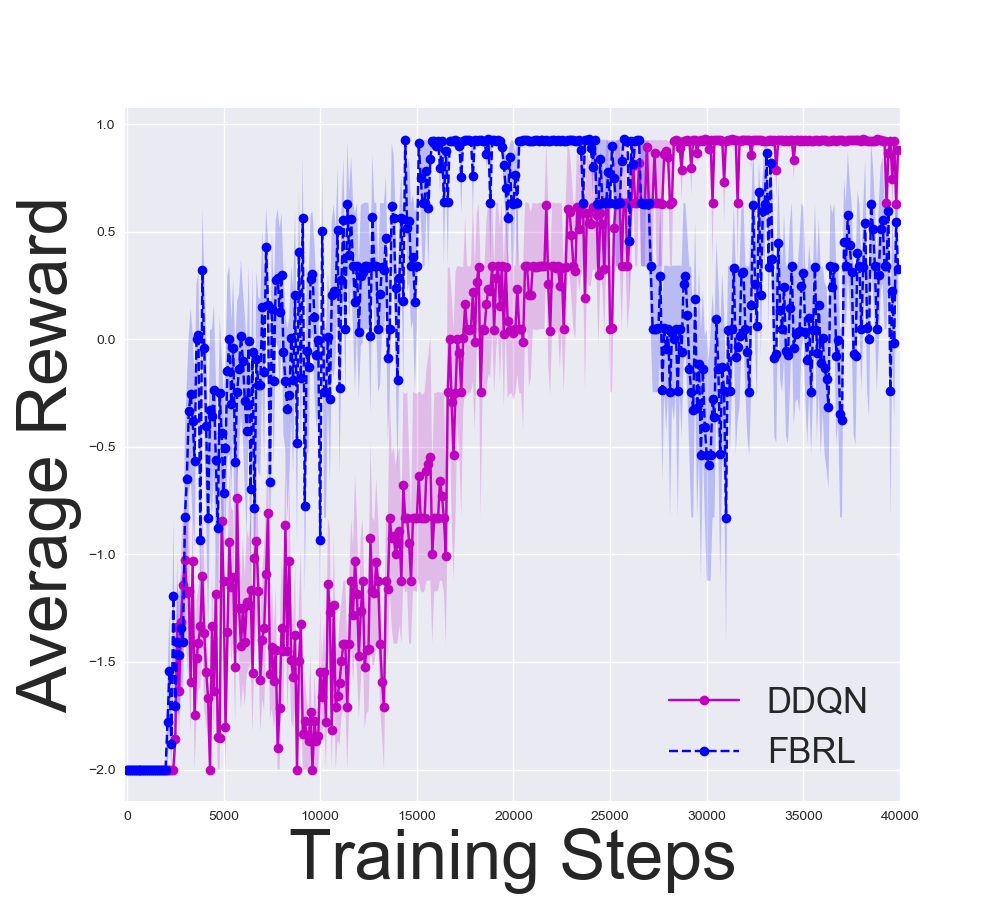}
  \end{subfigure}  
  \caption[]{Results for Towers of Hanoi where $n=2, 3$. We use a fixed horizon of $50,100$ steps, respectively. The results are averaged over $10$ trials. }  
  \label{fig:hanoi_results}
\end{figure}
The next environment we evaluate in is $n$-disc Towers of Hanoi. In this problem, the agent needs to move $n$ discs from the first to the third pillar, but it is only able to place a disc on top of another one if it is smaller than it. The actions are to move each disc to the first, second, or third pillar. It receives a reward of $1$ when all discs are in the third pillar and a step cost of $-.01$ per time-step. The inputs to the backward model are bit-strings indicating which pillars each disc is on. For example, the environment in Figure~\ref{fig:environments} has a representation of $[1,0,0,1,0,0]$ since the small disc is on the first pillar and the large disc is on the third pillar. The backward model predicts a distribution for each bit over possible $\Delta$ values: $P(\Delta=-1), P(\Delta=0), P(\Delta=1)$. The model architecture is a fully-connected network with $100$ outputs followed by RELU,  followed by another fully-connected network with $9n$ outputs, representing the distribution over each bit. For FBRL, we used $5$ steps of imagination with $3$ asynchronous streams.

Figure~\ref{fig:hanoi_results} shows the results for running Towers of Hanoi with a different number of discs. We again see an advantage for using FBRL as the goal gets further away. When we increase the number of discs, FBRL outperforms DDQN. We did find though that the performance of FBRL degraded for $3$ discs, which may be due to overfitting.

\section{Conclusion}
In this paper, we have introduced an approach for speeding up learning in problems with sparse rewards. We introduced FBRL, which takes imagined steps in reverse from the goal. We demonstrated that this approach can perform better than DDQN in Gridworld and Towers of Hanoi. There are many directions for extending this work. We were interested in evaluating a backward planner, but we could also train using both forward and backward imagination. Another improvement would be to improve the planning policy. We used a exploratory and greedy approach, but did not evaluate how to balance the two. We could also use prioritized sweeping~\citep{moore1993prioritized}, which chooses actions that lead to states with high TD-error. 

\section{Acknowledgements}
We thank Anoop Korattikara, Himanshu Sahni, Sergio Guadarrama, and Shixiang Gu for useful discussions and feedback about this work. 

\bibliography{references}

\begin{thebibliography}{14}
\providecommand{\natexlab}[1]{#1}
\providecommand{\url}[1]{\texttt{#1}}
\expandafter\ifx\csname urlstyle\endcsname\relax
  \providecommand{\doi}[1]{doi: #1}\else
  \providecommand{\doi}{doi: \begingroup \urlstyle{rm}\Url}\fi

\bibitem[Agrawal et~al.(2016)Agrawal, Nair, Abbeel, Malik, and
  Levine]{agrawal2016learning}
Agrawal, Pulkit, Nair, Ashvin~V, Abbeel, Pieter, Malik, Jitendra, and Levine,
  Sergey.
\newblock Learning to poke by poking: Experiential learning of intuitive
  physics.
\newblock In \emph{Advances in Neural Information Processing Systems}, pp.\
  5074--5082, 2016.

\bibitem[Baldassarre(2003)]{baldassarre2003forward}
Baldassarre, Gianluca.
\newblock Forward and bidirectional planning based on reinforcement learning
  and neural networks in a simulated robot.
\newblock In \emph{Anticipatory behavior in adaptive learning systems}, pp.\
  179--200. Springer, 2003.

\bibitem[Florensa et~al.(2017)Florensa, Held, Wulfmeier, and
  Abbeel]{florensa2017reverse}
Florensa, Carlos, Held, David, Wulfmeier, Markus, and Abbeel, Pieter.
\newblock Reverse curriculum generation for reinforcement learning.
\newblock \emph{arXiv preprint arXiv:1707.05300}, 2017.

\bibitem[Gu et~al.(2016)Gu, Lillicrap, Sutskever, and Levine]{gu2016continuous}
Gu, Shixiang, Lillicrap, Timothy, Sutskever, Ilya, and Levine, Sergey.
\newblock Continuous deep q-learning with model-based acceleration.
\newblock In \emph{International Conference on Machine Learning}, pp.\
  2829--2838, 2016.

\bibitem[Mnih et~al.(2015)Mnih, Kavukcuoglu, Silver, Rusu, Veness, Bellemare,
  Graves, Riedmiller, Fidjeland, Ostrovski, et~al.]{mnih2015human}
Mnih, Volodymyr, Kavukcuoglu, Koray, Silver, David, Rusu, Andrei~A, Veness,
  Joel, Bellemare, Marc~G, Graves, Alex, Riedmiller, Martin, Fidjeland,
  Andreas~K, Ostrovski, Georg, et~al.
\newblock Human-level control through deep reinforcement learning.
\newblock \emph{Nature}, 518\penalty0 (7540):\penalty0 529--533, 2015.

\bibitem[Moore \& Atkeson(1993)Moore and Atkeson]{moore1993prioritized}
Moore, Andrew~W and Atkeson, Christopher~G.
\newblock Prioritized sweeping: Reinforcement learning with less data and less
  time.
\newblock \emph{Machine learning}, 13\penalty0 (1):\penalty0 103--130, 1993.

\bibitem[Ng et~al.(1999)Ng, Harada, and Russell]{ng1999policy}
Ng, Andrew~Y, Harada, Daishi, and Russell, Stuart.
\newblock Policy invariance under reward transformations: Theory and
  application to reward shaping.
\newblock In \emph{ICML}, volume~99, pp.\  278--287, 1999.

\bibitem[Pathak et~al.(2017)Pathak, Agrawal, Efros, and
  Darrell]{pathak2017curiosity}
Pathak, Deepak, Agrawal, Pulkit, Efros, Alexei~A, and Darrell, Trevor.
\newblock Curiosity-driven exploration by self-supervised prediction.
\newblock In \emph{International Conference on Machine Learning (ICML)}, volume
  2017, 2017.

\bibitem[Schaul et~al.(2015)Schaul, Quan, Antonoglou, and
  Silver]{schaul2015prioritized}
Schaul, Tom, Quan, John, Antonoglou, Ioannis, and Silver, David.
\newblock Prioritized experience replay.
\newblock \emph{arXiv preprint arXiv:1511.05952}, 2015.

\bibitem[Sutton(1990)]{sutton1990integrated}
Sutton, Richard~S.
\newblock Integrated architectures for learning, planning, and reacting based
  on approximating dynamic programming.
\newblock In \emph{Proceedings of the seventh international conference on
  machine learning}, pp.\  216--224, 1990.

\bibitem[Sutton \& Barto(1998)Sutton and Barto]{sutton1998reinforcement}
Sutton, Richard~S and Barto, Andrew~G.
\newblock \emph{Reinforcement learning: An introduction}, volume~1.
\newblock MIT press Cambridge, 1998.

\bibitem[Van~Hasselt et~al.(2016)Van~Hasselt, Guez, and Silver]{van2016deep}
Van~Hasselt, Hado, Guez, Arthur, and Silver, David.
\newblock Deep reinforcement learning with double q-learning.
\newblock In \emph{AAAI}, volume~16, pp.\  2094--2100, 2016.

\bibitem[Weber et~al.(2017)Weber, Racani{\`e}re, Reichert, Buesing, Guez,
  Rezende, Badia, Vinyals, Heess, Li, et~al.]{weber2017imagination}
Weber, Th{\'e}ophane, Racani{\`e}re, S{\'e}bastien, Reichert, David~P, Buesing,
  Lars, Guez, Arthur, Rezende, Danilo~Jimenez, Badia, Adria~Puigdom{\`e}nech,
  Vinyals, Oriol, Heess, Nicolas, Li, Yujia, et~al.
\newblock Imagination-augmented agents for deep reinforcement learning.
\newblock \emph{arXiv preprint arXiv:1707.06203}, 2017.

\bibitem[Zang et~al.(2007)Zang, Irani, and Isbell~Jr]{zang2007horizon}
Zang, Peng, Irani, Arya, and Isbell~Jr, Charles~Lee.
\newblock Horizon-based value iteration.
\newblock Technical report, Georgia Institute of Technology, 2007.

\end{thebibliography}
\bibliographystyle{natbib}

\appendix
\section{Experimental setup}
For each experiment, we used a batch-size of $100$. The discount factor was $.99$. The exploration parameter $\epsilon$ was initialized to $1$ and decayed to $.1$. The replay memory had a size of $10000$ and we collected $10000$ initial samples before training DDQN. The architectures for the backwards models are described in the main text.

\subsection{Gridworld}
The learning rate for DDQN was $1e^{-3}$. The architecture for DDQN was a fully-connected network with $32$ outputs followed by RELU, followed by another fully-connected network with $4$ outputs, one for each action. We updated the target network every $100$ steps. FBRL had the same settings except we increased the learning rate to $5e^{-3}$.

\subsection{Towers of Hanoi}
The learning rate for DDQN was $5e^{-4}$. The architecture for DDQN was a fully-connected network with $32$ outputs followed by RELU, followed by another fully-connected network with $9$ outputs, one for each action. We updated the target network every $500$ steps. Like with Gridworld, we had the same architecture as DDQN, but we found we obtained better results when the learning rate was reduced to $1e^{-4}$. 

\end{document}